%% file: conference_101719.tex
\documentclass[conference]{IEEEtran}
\IEEEoverridecommandlockouts
\usepackage{cite}
\usepackage{amsmath,amssymb,amsfonts}
\usepackage{algorithmic}
\usepackage{graphicx}
\usepackage{textcomp}
\usepackage{xcolor}
\usepackage{booktabs}
\usepackage{multirow}
\usepackage[table]{xcolor}
\def\BibTeX{{\rm B\kern-.05em{\sc i\kern-.025em b}\kern-.08em
    T\kern-.1667em\lower.7ex\hbox{E}\kern-.125emX}}
\begin{document}

\title{SetFlow: Generating Structured Sets of Representations for Multiple Instance Learning\\
}

\author{
\IEEEauthorblockN{
Nikola Jovi\v{s}i\'c,
Milica \v{S}kipina,
Vanja \v{S}venda}
\IEEEauthorblockA{
Institute for Artificial Intelligence R\&D of Serbia\\
Novi Sad, Serbia\\
\{nikola.jovisic, milica.skipina, vanja.svenda\}@ivi.ac.rs}
}

\maketitle

\begin{abstract}
Data scarcity and weak supervision continue to limit the performance of machine learning models in many real-world applications, such as mammography, where Multiple Instance Learning (MIL) often offers the best formulation. While recent foundation models provide strong semantic representations out of the box, effective augmentation of such representations of MIL data remains limited, as existing methods operate at the instance level and fail to capture intra-bag dependencies. In this work, we introduce \textit{SetFlow}, a generative architecture that models entire MIL bags (i.e., sets) directly in the representation space. Our approach leverages the flow matching paradigm combined with a Set Transformer-inspired design, enabling it to handle permutation-invariant inputs while capturing interactions between instances within each bag. The model is conditioned on both class labels and input scale, allowing it to generate coherent and semantically consistent sets of representations. We evaluate SetFlow on a large-scale mammography benchmark using a state-of-the-art MIL-PF classification pipeline. The generated samples are shown to closely match the original data distribution and even improve downstream performance when used for augmentation. Furthermore, training on synthetic data alone shows competitive results, demonstrating the effectiveness of representation-space generative modeling for data-scarce and privacy-sensitive tasks.
\end{abstract}


\section{Introduction}
Although deep learning has achieved strong progress in a wide range of applications, its success in real-world settings remains constrained by data scarcity, label noise, and distributional complexity. These challenges are particularly pronounced in Multiple Instance Learning (MIL) \cite{dietterich1997solving,zaheer2017deep,ilse2018attention}, where each sample is represented as a bag (i.e., set) of instances, with supervision provided only at the bag level. This setting arises naturally in domains where fine-grained annotations are expensive or unavailable. One such example is mammography, where each examination consists of multiple high-resolution views, and labels are typically available only at the breast level, making it a challenging and clinically relevant MIL problem.

Recent advances in foundation models have enabled the extraction of rich semantic representations from images, leading to significant improvements in downstream tasks such as classification. However, despite these gains, the limited availability of labeled data continues to hinder generalization, especially in clinically diverse settings. Data augmentation has long been an attractive approach for addressing this issue, but conventional techniques (whether based on pixel-space transformations or instance-level generative modeling) fail to capture the complex structure of MIL data. In particular, they do not model the joint distribution of entire bags, nor do they preserve the intra-bag relationships.

\begin{figure}[]
\centering
\includegraphics[width=0.5\textwidth]{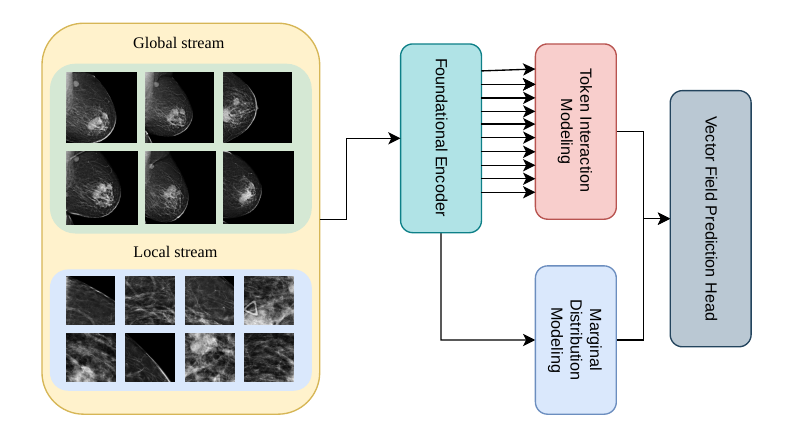}
\caption{\textbf{Overview of the proposed method}. Global (mammography views) and local (potential regions of interest in high resolution) streams are all individually encoded using a foundational encoder. Each instance model marginal instance distribution, while whole bags (as indicated by multiple arrows)  together capture the interaction of instances. Information from both streams are jointly leveraged to generate new sets of embeddings.
}
\label{fig:teaser}
\end{figure}

To address these limitations, we propose \textit{SetFlow}, an architecture designed to generate distributions of complete MIL bags directly in the representation space of foundation models. Our approach builds on the flow matching generative paradigm \cite{lipman2022flow}, and adapts it to set structured inputs. By leveraging a Set Transformer \cite{lee2019set}-inspired architecture, SetFlow captures both marginal instance distributions and instance interactions within a bag, as on Fig. \ref{fig:teaser}, while remaining computationally efficient for multiscale and variable-sized inputs.

We evaluate SetFlow in the context of mammography classification using the cutting-edge MIL-PF \cite{jovisic2026milpf} pipeline and demonstrate high quality of synthetic data, which can improve downstream performance when used for augmentation and even achieve competitive performance when used for training only by itself.

\section{Background} 

\subsection{Multiple Instance Learning (MIL).}
In Multiple Instance Learning (MIL) \cite{dietterich1997solving,zaheer2017deep,ilse2018attention}, the dataset $\mathcal{D}$ is composed of bags (sets) $\mathcal{S}_i$, as defined in Eq. \ref{eq:dataset}.

\begin{equation}
\resizebox{0.9\columnwidth}{!}{$
\mathcal{D} = \left\{ (\mathcal{S}_i, y_i) \right\}, 
\quad \text{where } 
\mathcal{S}_i = \{ I_i^{(n)} \}_{n=1}^{N_i} 
\text{ and } y_i \in \mathcal{L}
$}
\label{eq:dataset}
\end{equation}

Each bag $\mathcal{S}_i$ is a collection of instances $I_i^{(n)}$. The label $y_i \in \mathcal{L}$ is a bag-level label that applies to the entire bag $\mathcal{S}_i$; the labels for individual instances $I_i^{(n)}$ are unknown. 
The learning task is to map each bag $\mathcal{S}_i$ to its corresponding label $y_i$. 

\subsection{MIL-PF}

Multiple Instance Learning on Precomputed Features (MIL-PF) \cite{jovisic2026milpf} provides a structured classification pipeline for high-resolution, weakly labeled mammography. In this setting, each sample is represented as a bag of instances corresponding to multiple views and regions of a breast, with supervision provided only at the breast level. To account for this, MIL-PF maps images and their regions into a fixed semantic representation space using a frozen foundation encoder, enabling efficient downstream modeling without end-to-end fine-tuning.

The method represents each bag using two complementary streams: global embeddings that capture low-frequency anatomical context and local embeddings corresponding to candidate regions of interest with diagnostically relevant high-resolution signals. The resulting dataset consists of structured bags of $(global\ embeddings,\ local\ embeddings,\ label)$, where both streams are necessary for prediction, but are processed independently prior to aggregation. This structured embedding space provides a natural basis for generative modeling of entire MIL bags, which we exploit in our approach.

\subsection{Set Transformer}
Set Transformer \cite{lee2019set} is an architecture designed to address permutation invariant inputs such as MIL bags. It uses Induced Set Attention Blocks (ISAB) to reduce computational complexity of self-attention in classic Transformers \cite{vaswani2017attention}, making it well suited for data scarce domains such as mammography.

\subsection{Flow matching}

Flow matching \cite{lipman2022flow} is a generative paradigm for learning a continuous transport from a base distribution to the data distribution. Given a noise sample $x_0 \sim p_0$ and data sample $x_1 \sim p_{\text{data}}$, we construct a (linear) interpolation path (Eq. \ref{eq:fm_path}).
\begin{equation}
x_t = (1 - t)x_0 + t x_1, \quad t \in [0,1].
\label{eq:fm_path}
\end{equation}

A neural network $v_\theta(x,t)$ is trained to predict the velocity field along this path by minimizing the loss from the Eq. \ref{eq:loss}.
\begin{equation}
\mathcal{L}(\theta) =
\mathbb{E}_{t,x_0,x_1}
\left[
\left\| v_\theta(x_t,t) - (x_1 - x_0) \right\|^2
\right].
\label{eq:loss}
\end{equation}

Sampling is done by integrating the learned dynamics which transports noise to the data distribution (Eq. \ref{eq:velocity}).
\begin{equation}
\frac{dx}{dt} = v_\theta(x,t),
\label{eq:velocity}
\end{equation}

When paired with a transformer-style backbone, this formulation is compatible with DiT \cite{peebles2023scalable}-style generation, where sets of variable sizes can be generated conjointly.

\subsection{Representation space augmentation}
Synthetic data generation has long been used to improve downstream model performance by enriching training distributions. Earlier approaches ranged from classical augmentation techniques such as Vicinal Risk Minimization (VRM) \cite{chapelle2000vicinal} to deep generative models including Generative Adversarial Networks (GANs) \cite{goodfellow2020generative} and Variational Autoencoders (VAEs) \cite{diederik2019introduction}, which learn to generate new samples in the observation space. Latent Diffusion Models (LDMs) \cite{rombach2022high} improved scalability by generating data in a compressed latent domain rather than directly in the data space. Building on these developments, recent work explores generating synthetic data directly in semantic representation spaces produced by large foundation models, enabling better alignment with downstream tasks. Flow matching has recently emerged as a promising paradigm for modeling such representations \cite{zheng2025diffusion, gui2025adapting}.

Synthetic data generation in the representation space has been explored for high-resolution MIL medical imaging, with AugDiff \cite{boutaj2025controllable} and HistAug \cite{shao2024augdiff} as representative examples. However, these methods operate at the individual instance level by perturbing the original instances to preserve intra-bag consistency and, therefore, cannot model coherent bags as a whole.

\begin{figure}[]
\centering
\includegraphics[width=0.5\textwidth]{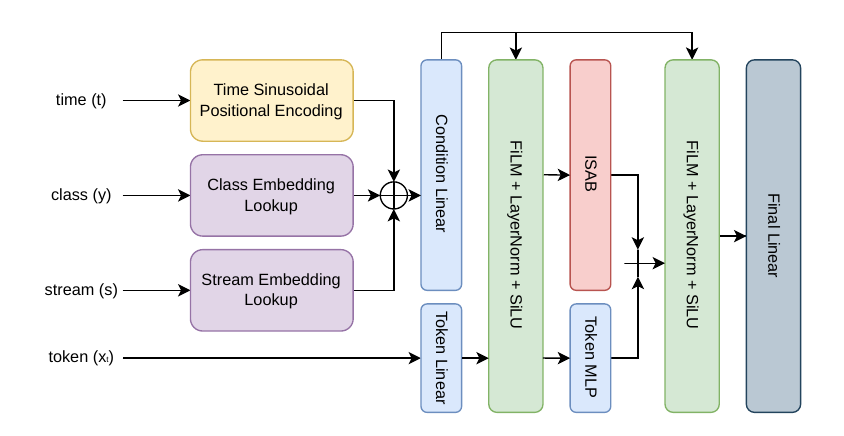}
\caption{\textbf{SetFlow architecture}.Time $t$, label $y$ and local/global stream identifier $s$ are embedded and concatenated to form the conditioning vector. The token is passed through a linear layer and conditioned on this vector before being processed by two branches: an MLP for deep marginal distribution modeling and an ISAB branch for capturing interactions between tokens. 
Finally, the outputs of both branches are summed, reconditioned, and passed to a final layer that predicts the vector field.}
\label{fig:architecture}
\end{figure}

\section{SetFlow}

SetFlow is an architecture that pioneers in modeling the distribution of whole MIL-suited bags (sets) of semantic representations. It generates these samples consistent with the MIL-PF pipeline, making it a \textit{generative counterpart of MIL-PF}. The architecture is shown in Fig. \ref{fig:architecture}.

\subsection{Model architecture}

For each token (global and local representation) in a bag, time $t$ is encoded using standard Transformer-style sinusoidal positional encoding\cite{vaswani2017attention}. The stream identifier $s$ (global/local) and label $y$ are encoded through an embedding lookup mechanism \cite{bengio2003neural}. These conditions are concatenated and passed through a linear layer to produce the conjoint condition $c$.

Each token (sample) $x_t$ is processed through it's own linear layer and conditioned on $c$ via FiLM \cite{perez2018film}, followed by LayerNorm \cite{ba2016layer} and SiLU \cite{ramachandran2017searching} activation. FiLM is used because the data is too scarce to support more complex mechanisms such as cross-attention. 

At this point in the network, each token has not yet:
\begin{itemize}
    \item passed through enough non-linear transformations to capture the marginal token distribution
    \item exchanged any information with other tokens in its bag
\end{itemize}

The first issue is solved by introducing a Multi-Layer Perceptron (MLP) branch that independently processes the token (\textit{Token MLP}). It consists of a cascade of linear layers followed by ELU \cite{clevert2016fast} activations. The second issue requires some type of attention mechanism. As there is insufficient data for quadratic complexity (self-attention), the other branch relies on the Set Transformer-inspired \textit{ISAB} block, which summarizes the whole bag into inducing points. The bag instances further query these points to update their state, implicitly exchanging information with other tokens in the bag. We suspect that even with more data this would be a correct inductive bias, as, unlike in the language domain, each-to-each interactions are not crucial in our case.

The outputs of these two branches are summed, additionally conditioned on the original $c$, and passed through LayerNorm, SiLU, and a final linear layer to predict the velocity field.

\subsection{PCA dimensionality reduction}
\label{subsec:pca}
The encoders used are trained on data distributions far broader than mammography, mapping inputs to a high-dimensional representation space. In this setting, mammograms are expected to lie on a lower-dimensional manifold, motivating dimensionality reduction of the embeddings. As shown in our results, Principal Component Analysis (PCA) is effective, retaining most of the important signal while significantly reducing the embedding size. This not only reduces computational cost, but also stabilizes flow matching training.

\subsection{Flow matching pipeline}
The complete training pipeline consists of reducing the dimensionality of foundational encoder embeddings, forming MIL-PF-style bags, and training SetFlow to model these bags using the flow matching procedure, which involves linear sampling and Mean Squared Error (MSE) loss.

\section{Experiments and Results} 
\subsection{Evaluation}
As SetFlow evaluation relies on MIL-PF, the experimental choices such as datasets and classification metrics follow the MIL-PF setup\cite{jovisic2026milpf}. We evaluate on EMBED, one of the largest public mammography datasets with over $0.5$M mammograms \cite{jeong2023emory}, and the widely used VinDr-Mammo dataset \cite{nguyen2023vindr}, both representing highly diverse real-world clinical scenarios  \cite{woo2025subgroup}.  As a classification pipeline we use the attention-based MIL-PF with DINOv2 ViT Giant \cite{oquab2023dinov2} and MedSigLIP from MedGemma \cite{sellergren2025medgemma} as backbones. Positive exam cases use BI-RADS 4-6, while negative cases use BI-RADS 1, following the MIL-PF narrative.

\subsection{Implementation details}
The model operates on input embeddings of dimension 128, projected to a hidden dimension of 512. Token representations are further projected to dimension 32 before applying the ISAB block with 4 inducing points, then mapped back to the hidden dimension. Conditioning uses 16-dimensional embeddings for time, class label, and stream, concatenated and projected into a joint conditioning vector. The final output is mapped back to the original embedding dimension.

The preprocessing pipeline is taken directly from MIL-PF. SetFlow is trained using the ADAM optimizer with a learning rate of 1e-4 on whole-dataset batches, as in MIL-PF. Training stops when no further improvement is observed in the internal FID (see Subsec. \ref{subsec:fid}), which is evaluated every 20k iterations. Integration is performed using Runge-Kutta (RK2) method, with 200 steps at inference. The ratio of local to global instances inside the bag is sampled from a Gaussian distribution estimated from the original datasets. The whole pipeline is implemented in PyTorch.

\subsection{PCA dimensionality reduction}

Before training SetFlow, the impact of PCA dimensionality reduction on area under receiver operator curve (AUC) of MIL-PF classifier is measured. As shown in Tab. \ref{tab:dim_reduction}, reducing to $128$ from the original $1536$ (DINOv2 ViT Giant) and $1152$ (MedSigLIP) preserves performance while significantly reducing computation cost. Further compression degrades performance, so all SetFlow experiments use the $128$ configuration.

\input{tables/pca}

\subsection{Original/synthetic distribution comparison}
\label{subsec:fid}

Unlike with images in pixel space, no visual evaluation of the quality of semantic representations is possible. Instead, we measure the similarity between the synthetic and original data using an adaptation of the Fréchet Inception Distance (FID) \cite{heusel2017gans}. As we are already operating in the latent domain, embeddings are compared directly, so the use of InceptionNetV3 \cite{szegedy2016rethinking} from the classic FID definition is unnecessary.

For both the original and synthetic data, we compute FID:
\begin{itemize}
    \item Internal: a similarity of random divisions of the set
    \item Interstream: a similarity of local and global tokens 
    \item Interclass: a similarity of positively and negatively labeled instances (only for global stream, for simplicity and noise reduction, as label signal is sparse in the local stream).
\end{itemize}

We also measure the similarity between the synthetic examples and the originals by directly comparing distributions.

As shown in Tab. \ref{tab:fid}, FID values are generally low compared to what is typically found in the literature. This is attributed to the low variance of the dataset (see Subsec. \ref{subsec:pca}), which leads to small FID values even after applying PCA. Still, significant drops in "w.r.t original" FID during training and plateauing at the very low values from the table indicate successful training.

Furthermore, comparing the relative differences for internal, interstream, and interclass values across the two corpora of data shows that the order of magnitude is preserved and the values are quite similar. We interpret this as a positive sign that our conditioning mechanism effectively encodes label and stream information.

\input{tables/fid}

\subsection{Nearest neighbour analysis}

One key concern when generating synthetic data is avoiding replicating examples from the original dataset and collapsing the generated data into low diversity. To assess SetFlow with regard to this, Euclidean distances between nearest neighbors are computed (see Tab. \ref{tab:nn}). Synthetic data exhibits a greater spread than the original data, indicating no signs of collapse. Additionally, the nearest original neighbor for a synthetic example (and vice versa) is farther away in space than the original is from its nearest original, suggesting that samples are not replicated. These results suggest that there is potential for modeling the original distribution more tightly, which we intend to explore in our future work.

\input{tables/nn}

\subsection{Classifier performance analysis}
MIL-PF is trained on original, combined, and synthetic datasets. Combined datasets include original data, enriched with 20\% additional synthetic examples per class (training set only). Synthetic datasets are generated by sampling as many positive and negative examples as there are in the original datasets and using only those examples.

Tab. \ref{tab:classification} shows values of AUC (for class separability), balanced accuracy (bACC; to fairly include performance on minority class), and specificity at sensitivity of 90\% (Spec@Sens=0.9; to show performance in clinically important scenarios, where the sensitivity threshold must be high).

A slight improvement in some metrics is present when using combined data for the EMBED dataset, while the synthetic-only data alone yield surprisingly competitive AUC and bACC results for the MedSigLIP encoder on the VinDr dataset. These results suggest that synthetic data align well with the original data and can even slightly contribute to the discriminative value. We consider this significant, especially considering the amount of noise that is introduced in the generative modeling by the sparsity of signal-carrying instances in the local stream.

\input{tables/classifier}







\section{Conclusion}

We introduced SetFlow, an efficient architecture for modeling the distribution of Multiple Instance Learning (MIL) bags of mammography images using flow matching. Our findings demonstrate that such a task is feasible, with synthetic examples aligning well with the original data and correctly capturing class information, scale, and intra-bag consistency. These findings open several directions for future research, including extending the approach to other domains, refining the SetFlow architecture and its flow matching procedure, and further improving the fidelity and usefulness of generated samples for data augmentation and downstream performance.

\section{Acknowledgements}

We thank Alexandros Graikos (Stony Brook University, New York) for insightful discussions and guidance that motivated our work on representation-space modeling.

\bibliographystyle{IEEEtran}
\bibliography{references}

\end{document}

%% file: tables/pca.tex
\begin{table}[htpb]
\centering
\caption{PCA impact on classification performance (AUC)}

\begin{tabular}{cccccc}
\toprule

\textbf{Backbone} & \textbf{Dataset} & \textbf{No reduction} & \textbf{128} & \textbf{32} & \textbf{8}\\ 
\midrule

MedSigLIP & EMBED & 0.914 & 0.904 & 0.878 & 0.819\\
MedSigLIP & VinDr & 0.911 & 0.909 & 0.900 & 0.827\\
DINOv2 & EMBED & 0.916 & 0.896 & 0.856 & 0.782\\
DINOv2 & VinDr & 0.894 & 0.885 & 0.824 & 0.666\\

\bottomrule
\end{tabular}

\label{tab:dim_reduction}
\end{table}

%% file: tables/fid.tex
\begin{table}[!ht]
\centering
\caption{FID-based comparison of original and synthetic distributions}
{\resizebox{\columnwidth}{!}{
\begin{tabular}{llcccc}
\toprule
& & \multicolumn{2}{c}{\textbf{EMBED}} & \multicolumn{2}{c}{\textbf{VinDr}} \\
\cmidrule(lr){3-4}\cmidrule(lr){5-6}
\textbf{Model} & \textbf{FID} & \textbf{Original} & \textbf{Synthetic} & \textbf{Original} & \textbf{Synthetic} \\
\midrule
\multirow{4}{*}{MedSigLIP}
& Internal & 8e-5 & 8e-5 & 2e-4 & 7e-5 \\
&  Interstream & 0.39 & 0.23 & 0.37 & 0.22 \\
& Interclass & 0.015 & 0.002 & 0.017 & 0.005 \\
&  W.r.t. original & -- & 0.07 & -- & 0.08 \\
\midrule
\multirow{4}{*}{DINOv2}
& Internal & 9e-5 & 7e-5 & 3e-4 & 5e-5 \\
& Interstream & 0.42 & 0.39 & 0.15 & 0.08 \\
& Interclass & 0.006 & 0.002 & 0.019 & 0.001 \\
&  W.r.t. original & -- & 0.20 & -- & 0.03 \\
\bottomrule
\end{tabular}
}}
\label{tab:fid}
\end{table}

%% file: tables/nn.tex
\begin{table}[!ht]
\centering
\caption{Euclidian nearest-neighbour evaluation for synthetic and original data}

{\resizebox{\columnwidth}{!}{
\begin{tabular}{llcc}
\toprule
\textbf{Model} & \textbf{Metric} & \textbf{EMBED} & \textbf{VinDr} \\

\midrule

\multirow{4}{*}{MedSigLIP}
& Internal (original) & 0.26 & 0.26 \\
& Internal (synthetic) & 0.35 & 0.33 \\
& Original $\rightarrow$ Synthetic & 0.34 & 0.32 \\
& Synthetic $\rightarrow$ Original & 0.35 & 0.34 \\

\midrule

\multirow{4}{*}{DINOv2}
& Internal (original) & 0.19 & 0.14 \\
& Internal (synthetic) & 0.29 & 0.18 \\
& Original $\rightarrow$ Synthetic & 0.26 & 0.19 \\
& Synthetic $\rightarrow$ Original & 0.39 & 0.18 \\

\bottomrule
\end{tabular}
}}

\label{tab:nn}
\end{table}

%% file: tables/classifier.tex
 \begin{table}[!ht]
  \centering
  \caption{Classification performance for original, synthetic and combined data}
  \resizebox{\columnwidth}{!}{
  \begin{tabular}{llcccccc}
  \toprule
   &  & \multicolumn{3}{c}{EMBED} & \multicolumn{3}{c}{VinDr} \\
  \cmidrule(lr){3-5} \cmidrule(lr){6-8}
  \textbf{Model} & \textbf{Metric}
  & Orig. & Comb. & Synth.
  & Orig. & Comb. & Synth. \\
  \midrule

  \multirow{3}{*}{MedSigLIP}
  & AUC & 0.904 & 0.904 & 0.774 & 0.909 & 0.857 & 0.855 \\
  &  bACC & 0.841 & 0.835 & 0.698 & 0.851 & 0.806 & 0.789 \\
  & Spec@Sens=0.9 & 0.698 & 0.728 & 0.384 & 0.681 & 0.653 & 0.283 \\

  \midrule

  \multirow{3}{*}{DINOv2}
  & AUC & 0.896 & 0.900 & 0.767 & 0.885 & 0.858 & 0.764 \\
  &  bACC & 0.833 & 0.841 & 0.699 & 0.829 & 0.797 & 0.638 \\
  & Spec@Sens=0.9 & 0.657 & 0.663 & 0.322 & 0.733 & 0.673 & 0.317 \\

  \bottomrule
  \end{tabular}}
  \label{tab:classification}
  \end{table}